\documentclass[10pt,journal,compsoc]{IEEEtran}

\usepackage{lipsum}

\usepackage{endnotes}

\usepackage{enumerate}

\usepackage[sort&compress,comma,square,numbers]{natbib}

\usepackage{amsfonts,amsmath,amssymb}
\usepackage{bbm}
\usepackage{xspace}

\usepackage{multicol}
\usepackage{enumitem}

\usepackage{hhline}

\usepackage{comment}
\usepackage{todonotes}
\RequirePackage{ulem}

\usepackage{booktabs}
\usepackage{fullpage}

\usepackage{algorithm}
\usepackage{algpseudocode}

\usepackage[super]{nth}
\usepackage{mdframed}
\mdfdefinestyle{MyFrame}{%
    outerlinewidth=6pt,
    roundcorner=20pt,
    innertopmargin=\baselineskip,
    innerbottommargin=\baselineskip,
    innerrightmargin=20pt,
    innerleftmargin=20pt 
    }

\newcommand{\argmin}{\operatornamewithlimits{argmin}}

\providecommand{\mc}[1]{\mathcal{#1}}

\providecommand{\mbb}[1]{\mathbb{#1}}

\newcommand{\iid}{\overset{iid}{\sim}}

\usepackage{hyperref}

\ifCLASSOPTIONcompsoc
  \usepackage[nocompress]{cite}
\else
  \usepackage{cite}
\fi
\usepackage{subcaption}

\ifCLASSINFOpdf
\else
\fi
\begin{document}
%
\title{Leveraging semantically similar queries for ranking via combining representations}
%
%
%
%

\author{Hayden~S.~Helm$^{1,2,\dagger}$, Marah Abdin$^{1}$,
Benjamin D. Pedigo$^{1,2}$,
Shweti~Mahajan$^{1}$, 
Vince Lyzinski$^{3}$,
Youngser~Park$^{2}$,
Amitabh Basu$^{2}$,
Piali~Choudhury$^{1}$,
Christopher~M.~White$^{1}$,
Weiwei~Yang$^{1}$,
Carey~E.~Priebe$^{2, \dagger}$ 
\thanks{$^{1} $ Microsoft Research}
\thanks{$^{2} $ Johns Hopkins University}
\thanks{$^{3} $ University of Maryland, College Park}
\thanks{$^{\dagger}$ corresponding authors: haydenshelm@gmail.com \& cep@jhu.edu}
}
\IEEEtitleabstractindextext{%
\begin{abstract}
In modern ranking problems, different and disparate representations of the items to be ranked are often available. It is sensible, then, to try to combine these representations to improve ranking. Indeed, learning to rank via combining representations is both principled and practical for learning a ranking function for a particular query.  
In extremely data-scarce settings, however, the amount of labeled data available for a particular query can lead to a highly variable and ineffective ranking function. 
One way to mitigate the effect of the small amount of data is to leverage information from semantically similar queries. 
Indeed, as we demonstrate in simulation settings and real data examples, when semantically similar queries are available it is possible to gainfully use them when ranking with respect to a particular query. 
We describe and explore this phenomenon in the context of the bias-variance trade off and apply it to the data-scarce settings of a Bing navigational graph and the \textit{Drosophila} larva connectome.
\end{abstract}

}

\maketitle


%
\IEEEpeerreviewmaketitle

\ifCLASSOPTIONcompsoc
\IEEEraisesectionheading{\section{Introduction}\label{sec:introduction}}
Ranking objects relative to a given query is of general interest, with applications ranging from improving search engines \citep{10.1007/s10791-009-9123-y,  pmlr-v14-chapelle11a}  to identifying potential human traffickers \citep{fishkind2015vertex}. Typically when ranking objects there is a domain specific notion of similarity such that the objects-to-be-ranked, or nomination objects, can be sorted relative to their closeness to the query. In other settings the definition of similar is poorly defined and needs to be learned using supervisory information or data. Using machine learning algorithms out-of-the-box to learn the relationship between the query and the nomination objects, however, often requires a significant amount of data. 

One way to mitigate the need for a large amount of data is to leverage pre-existing models or representations. That is, rather than learning a model or representation of the data for a particular query from scratch, simply update models already trained or representations already learned that are presumed to be useful for downstream inference. This paradigm for learning has shown success in resource-scarce settings such as transfer learning \citep{pan2009survey}, domain adaptation \citep{wang2018deep}, and continual learning \citep{vogelstein2020general}.

In the ranking setting, Helm et al \citep{Helm2020LearningTR} demonstrate the effectiveness of this paradigm by learning an optimal linear combination of pre-existing dissimilarities given a set of nomination objects known to be similar to the query. The learned linear combination of pre-existing dissimilarities is tailored specifically to the particular query and does not necessarily generalize well to other queries. In this paper we introduce an extension to the method evaluated in Helm et al to learn an optimal linear combination of pre-existing distances for a set of semantically similar queries in the context of vertex nomination \citep{JMLR:v20:18-048}. We then discuss and demonstrate its utility in principled simulation settings and practical real data examples.

We consider a similar setting to the one described in Helm et al \citep{Helm2020LearningTR}. Recall that in their setting we are given a query $ q $ and asked to rank a set of nomination objects $ \mc{N} $ relative to $ q $. The ``correct" notion of similarity for which to rank the elements of $ \mc{N} $ is unknown and instead the definition of similar needs to be inferred, or learned, from a set of nomination objects $ S \subseteq \mc{N} $ known to be similar to $ q $ and a set of dissimilarity measures defined on the query and nomination objects. 
The set $ S $ implicitly defines a useful ranking and can be used in combination with the set of dissimilarities to improve rank-based inference \textit{without the need to train complex models from scratch}. Their setting is similar to multi-view learning using only positive and unlabeled examples \citep{pmlr-v25-zhou12} with the twist that the original feature vectors are not available and, instead, only distances or dissimilarities are available.

The setting considered herein is a natural extension of the setting described in Helm et al. In particular, we have access to $ K $ $ (q, S) $ pairs and a set of dissimilarities defined on each of the queries and the entire set of nomination objects. We assume that the queries are ``semantically similar" -- or, that the expected value of the definition of similar implicitly defined the corresponding $ S $ sets is the same. As we will demonstrate in simulations and in real data examples, it is possible to gainfully use a distance learned from the additional $ (q,S) $ pairs for ranking relative to a particular query of interest.

\else
\section*{Introduction}
\label{sec:introduction}
\fi
\subsection*{Ranking, learning to rank, and PU Learning}

In the ranking setting there exists a set of queries $ \mc{Q} $ and a set of nomination objects $ \mc{N} $ and the goal, given an element $ q \in \mc{Q} $, is to rank the elements of $ \mc{N} $ such that the elements of $ \mc{N} $ most similar or relevant to $ q $ are near the top of a ranked list. We let $ \mc{H} $ be the set of ranking functions defined on $ \mc{Q} \times \mc{N} $. Or, an element of $ \mc{H} = \{h: \mc{Q} \times \mc{N} \to \{1, \hdots, |\mc{N}|\}\} $ takes as input a query and a nomination object and outputs an integer, interpreted to be a rank of how similar the nomination object is to the query, from 1 to the size of the set of nomination objects. As in \citep{Helm2020LearningTR}, we ignore the possibility of ranking ties for expediency. See Appendix B.1 of \citep{JMLR:v20:18-048} for a discussion.

An intuitive way to rank the elements of $ \mc{N} $ relative to $ q $ is to use a dissimilarity measure $ d $ defined on the space $ \mc{Q} \times \mc{N} $. That is, $ d: \mc{Q} \times \mc{N} \to \mbb{R} $ is such that for a fixed $ q $ small values of $ d $ correspond to elements of $ \mc{N} $ similar or relevant to $ q $. The ranking of $ \mc{N} $ induced by $ d $ is just the sorted dissimilarities. We let $ h^{d} \in \mc{H} $ be the ranking function induced by $ d $ (i.e. $ h^{d}(\argmin_{v \in \mc{N}} d(q, v)) = 1 $).

In the learning to rank \citep{liu2011learning} and related \citep{conte2004thirty} settings, there exists a set of supervisory information that enables the learning of effective ranking functions. Supervisory information can come in a myriad of forms -- as a set of (feature vector, ordinal) pairs or as a set of nomination objects known to be similar to $ q $, for two examples. 

If the supervisory information comes in the form of a set of nomination objects known to be similar to $ q $, the learning to rank problem is similar to learning from Positive and Unlabeled (PU) examples \citep{Bekker_2020}. In PU Learning, researchers have access to positive examples of a class of interest and a set of unlabeled examples unknown to be either positive or negative and the goal is to correctly identify elements of the class of interest in the set of unlabeled data. Under particular sampling assumptions, standard classifiers such as Support Vector Machines, Random Forests, and Naive Bayes can be adapted to take as input only positive and unlabeled data \citep{10.1145/1401890.1401920}. For a given unlabeled observation, these classifiers output a posterior probability of being in the class of interest. The posteriors can be sorted to produce a ranked list.

In general, the goal in learning to rank is to use a learning algorithm that outputs  a (potentially random) ranking function $ h $ that performs well relative to an evaluation criterion, such as Mean Reciprocal Rank (MRR). The MRR of $ h \in \mc{H} $ for a query $ q \in \mc{Q} $ and evaluation set $ \mc{N}' \subseteq \mc{N} $ is the average of the multiplicative inverse (or reciprocal) of the $ h(q, s) $ for $ s \in \mc{N}' $. That is,
\begin{align*}
    \text{MRR}(h, q, \mc{N}') = \frac{1}{|\mc{N}'|}\sum_{s \in \mc{N}'} \frac{1}{h(q, s)}.
\end{align*} Notably, the optimal MRR for a given $ \mc{N}' $ decreases as the size of $ \mc{N}' $ increases. The ranking function $ h $ is preferred to the ranking function $ h' $ for $ \mc{N}' $ if $ \text{MRR}(h, q, \mc{N}') > \text{MRR}(h', q, \mc{N}') $. Other evaluation criteria include Mean Average Precision and Recall at $ k $ \citep{radev2002evaluating}.

In the simulation settings below we evaluate various ranking functions using \textit{normalized} mean reciprocal rank. Normalized mean reciprocal is the ratio of the mean reciprocal rank of a ranking function $ h $ with respect to some query $ q $ and an evaluation set $ \mc{N}' $ and the mean reciprocal rank of an optimal ranking function $ h^{*} $ that places the evaluation set $ \mc{N}' $ at the top of the rank list with respect to the query $ q $:
\begin{align*}
    \text{Normalized MRR}(h, q, \mc{N}') = \frac{\text{MRR}(h, q, \mc{N}')}{\text{MRR}(h^{*}, q, \mc{N}')}.
\end{align*} Normalized mean reciprocal rank partially mitigates the effect of varying sizes of the evaluation set $ \mc{N}' $. 

\subsection*{Vertex nomination} Vertex nomination \citep{JMLR:v20:18-048, coppersmith2014vertex} is an instance of the ranking problem where both the query and the nomination objects are vertices in a network. Recall that a graph or network $ G = (V, E) $ is a set of vertices $ V $ and a (potentially asymmetric) pairwise relationship defined on the vertices. For example, a connectome \citep{eichler2017complete} is a network whose vertices are neurons and the pairwise relationship is an indicator of whether or not two neurons share a synapse. We let $ V = \{1, \hdots, n\} $ and $ \mc{G}_{n} $ be the set of networks with $ n $ vertices.
Typically in vertex nomination the set of query objects $ \mc{Q} $ is a single vertex $ \{v^{*}\} $ with $ v^{*} \in V $ and the set of nomination objects is the set of vertices not including the query $ V \setminus \{v^{*}\} $.

Recently proposed methods for vertex nomination \citep{marchette2011vertex, coppersmith2012vertex, sun2012comparison, suwan2015bayesian, fishkind2015vertex, agterberg2019vertex, yoder2020vertex} have been quite successful in a variety of settings when the definition of similar is defined explicitly by a domain expert. Approaches are mainly combinatorial or spectral. 

\subsection*{Multiple representations and dissimilarities}
It is often the case that multiple representations of the query and nomination objects are available -- from simple scaling, shifting and  projecting to considering entirely different sets of features -- and that the optimal representation for inference is not known. Further, even in the case of a single representation of data, different dissimilarities (e.g. Euclidean vs Mahalanobis vs Cosine) may be deployed for ranking. This is particularly relevant for vertex nomination when considering the various \textit{embedding} methods \citep{von2007tutorial,grover2016node2vec,tsitsulin2018verse, hamilton2017inductive, doi:10.1080/01621459.2012.699795,  qiu2019netsmf}.

A graph or network embedding method takes as input a graph $ G = (V,E) $ and outputs a vector representation for each vertex in $ V $. We let $ T^{j}: \mc{G}_{n} \to \{\mbb{R}^{m_{j}}\}^{n} $ be the embedding method indexed by $ j $ and $ T^{j}(G)_{i} $ be the vector representation of vertex $ i $ resulting from using the $ jth$ embedding method. Note that with each embedding method typically comes a dissimilarity defined on the embedded space. We let $ d^{j}: \mbb{R}^{m_{j}} \times \mbb{R}^{m_{j}} \to \mbb{R} $ be the dissimilarity associated with $ T^{j} $. For ease of notation we often have $ d^{j} $ take elements of $ V $ as input.

Further, different types of relationships can be measured on the same set of vertices: $ G_{1} = (V, E_{1}), G_{2} = (V, E_{2}), \hdots, G_{J} = (V, E_{J}) $. The choice of edge type to use can similarly impact downstream inference and we think of this leading to a different set of representations of the vertices.

In general, different embedding methods, different dissimilarities and different edge types defined on the same set of objects can induce different ranking functions, potentially impacting downstream inference \citep{Helm2020LearningTR}. Hence, embedding method and dissimilarity selection is a critical step when ranking vertices within a network.

\subsection*{An Integer Linear Program: Single Query}

Supervision of the form of vertices known to be ``similar" to the query can be used to choose, or learn, a dissimilarity to deploy for ranking. 

We borrow heavily from the problem setting and method discussed in Helm et al and thus describe their problem and proposed integer linear program in detail. 

Let $ V = \{v_{1}, \hdots, v_{n}\} $ be a set of items with $ v^{*} := v_{1} $ without loss of generality. Suppose there exists $ J $ distinct dissimilarity measures $ d^{j}: V \times V \to \mbb{R} $ for $ j \in \{1, .., J\} $ and that we have access to the dissimilarities between $ v^{*} $ and all of the other items: $ \{\{d^{j}(v^{*}, v_{i})\}_{i=2}^{n}\}_{j=1}^{J} $. Further suppose that we have access to a set $ S \subseteq \{v_{2},  \hdots, v_{n}\} $ of items known to be similar to $ v^{*} $. We assume the elements of the candidate set $ C = V \setminus \{\{v^{*}\} \cap S\} $ could be either similar or not similar to $ v^{*} $. The goal, as in learning to rank, is to produce a ranking function $ h $ that maps the unknown positive example to small values.

Since they make no assumptions on the dissimilarities or the elements of S, there are no theoretical properties of either to attempt to exploit. Instead, it is assumed that a ranking function that minimizes the maximum ranking of an element of $ S $ will similarly map positive examples in the set of unlabeled examples to small values. For example, if the set of ranking functions was restricted to $ \{h^{d^{j}}\}_{j=1}^{J} $ then we'd use the $ h^{d^{j}} $ such that $$ \max_{s \in S} h^{d^{j}}(s) $$ is minimized.

To enrich the set of hypotheses we can consider convex combinations of the dissimilarities. To do this, we introduce a set of nonnegative weights $ \alpha_{1}, \hdots, \alpha_{J} $ such that $ \sum _{j=1}^{J} \alpha_{j} = 1 $. Letting $ \vec{\alpha} = (\alpha_{1}, \hdots, \alpha_{J}) $, $ \Delta^{J-1} $ be the $ J $ dimensional simplex, and $ h^{\vec{\alpha}} $ the ranking function induced by $ \sum_{j=1}^{J} \alpha_{j} d^{j} $ we solve the following optimization problem:
\begin{equation}\label{eq:ilp}
    \min_{\vec{\alpha}} \max_{s \in S} h^{\vec{\alpha}}(s).
\end{equation}

The optimization problem \eqref{eq:ilp} can be solved using a mixed Integer Linear Program (ILP). To set up the ILP we introduce the indicator variables $ x^{\vec{\alpha}}_{v} $ for all $ v \in C $ where
\begin{align*}
    x^{\vec{\alpha}}_{v} = \begin{cases} 0 \text{ if }  h(v) > h(s) \quad \forall s \in S \\
    1 \text{ o.w.}
    \end{cases}
\end{align*} The ILP objective can then be defined as the sum of the indicator variables
\begin{align*}
    \min_{\vec{\alpha}} \sum_{v \in C} x_{v}
\end{align*} with the linear constraints 
\begin{align*}
    \sum_{j=1}^{J} \alpha_{j} d^{j}(v^{*}, s) \leq \sum_{j=1}^{J} \alpha_{j} d^{j}(v^{*}, v) + M\cdot x^{\vec{\alpha}}_{v} \\ 
\;\; \forall (s, v) \in (S, C)
\end{align*} to ensure that $ x_{v} = 0 $ only when $ v \in C $ is ranked below all of the elements of $ S $ and where $ M = \max_{i} \{d^{j}(v^{*}, v_{i})\}_{i=2}^{n} $.

Using the ILP to produce a ranking function significantly improved upon the performance, as measured by MRR, of the algorithm that selects a ranking function amongst the best $ \{h^{d^{j}}\}_{j=1}^{J} $ in applications including ranking voxels for a particular brain region, ranking input neurons in the mushroom body of the \textit{Drosophila}, and ranking phones and devices for Bing.

There are two main drawbacks of the proposed ILP. Firstly, the learned distance is tailored specifically to the vertex of interest and, in general, will not generalize to queries whose definition of similar is (even slightly) different. This implies that a new distance should be learned for every query, even if the queries are semantically similar. Further, due to the computational complexity of the ILP, learning a distance for each possible query may be untenable.

Secondly, for the ILP to be effective, the number positive examples needed to learn an effective ranking function is too large to be practical. For example, when ranking objects from the set of searchable objects in Bing it is infeasible to hand label multiple objects known to be similar to each query. In such cases, leveraging semantically similar queries and their corresponding sets of objects known to be similar to them is necessary to be able to learn an effective ranking scheme.

In the next section we introduce a distribution on $ (q, S) $ pairs, thus motivating the development and employment of an integer linear program that learns a distance using a collection of $ (q, S) $ pairs. 

\begin{figure*}[t!]
    \captionsetup[subfigure]{justification=centering}
    \begin{subfigure}{\textwidth}
    \centering
    \includegraphics[width=0.45\linewidth]{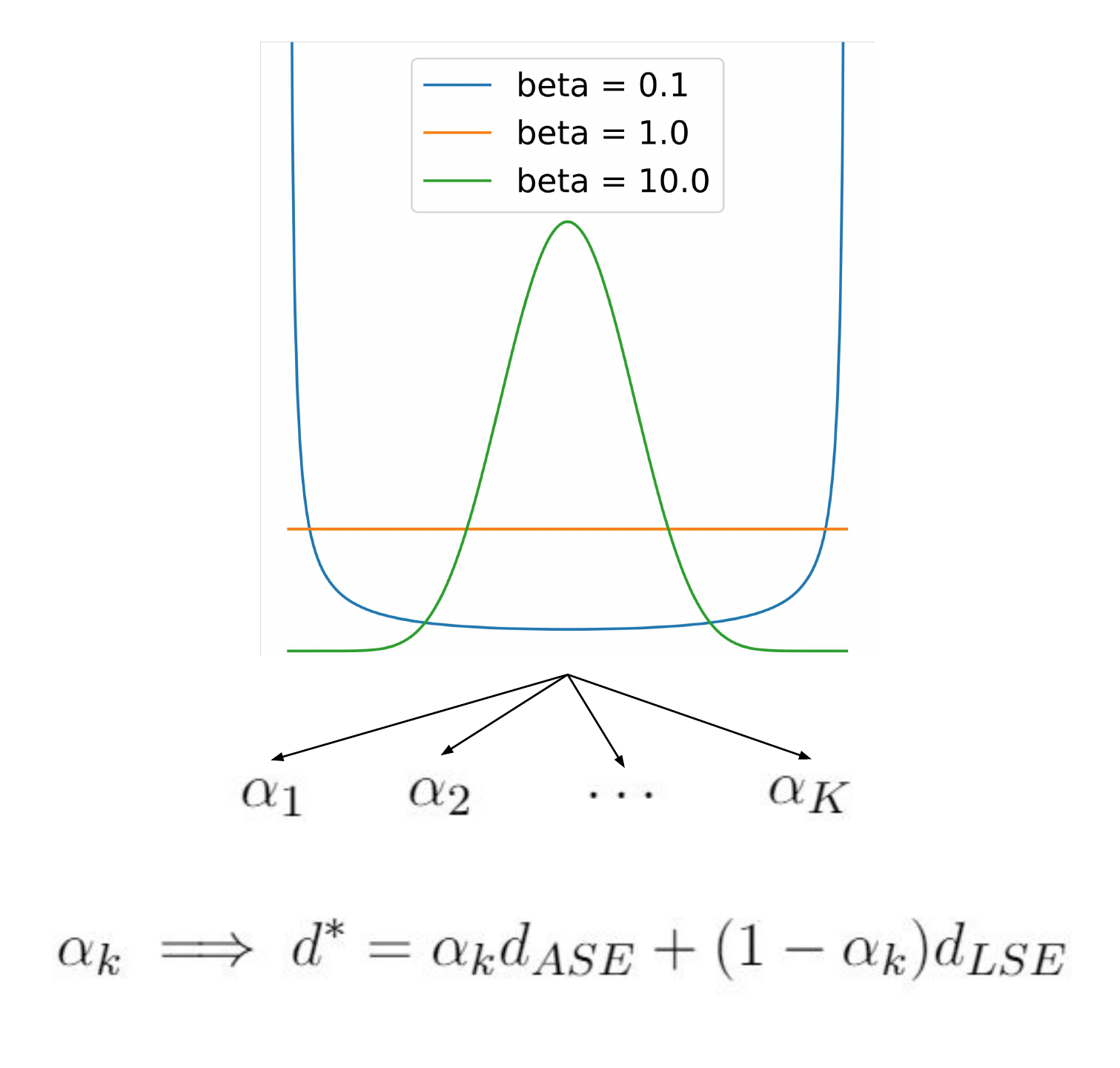} \\
    \end{subfigure}
    \begin{subfigure}{.33\textwidth}
          \centering
          \includegraphics[width=.9\linewidth]{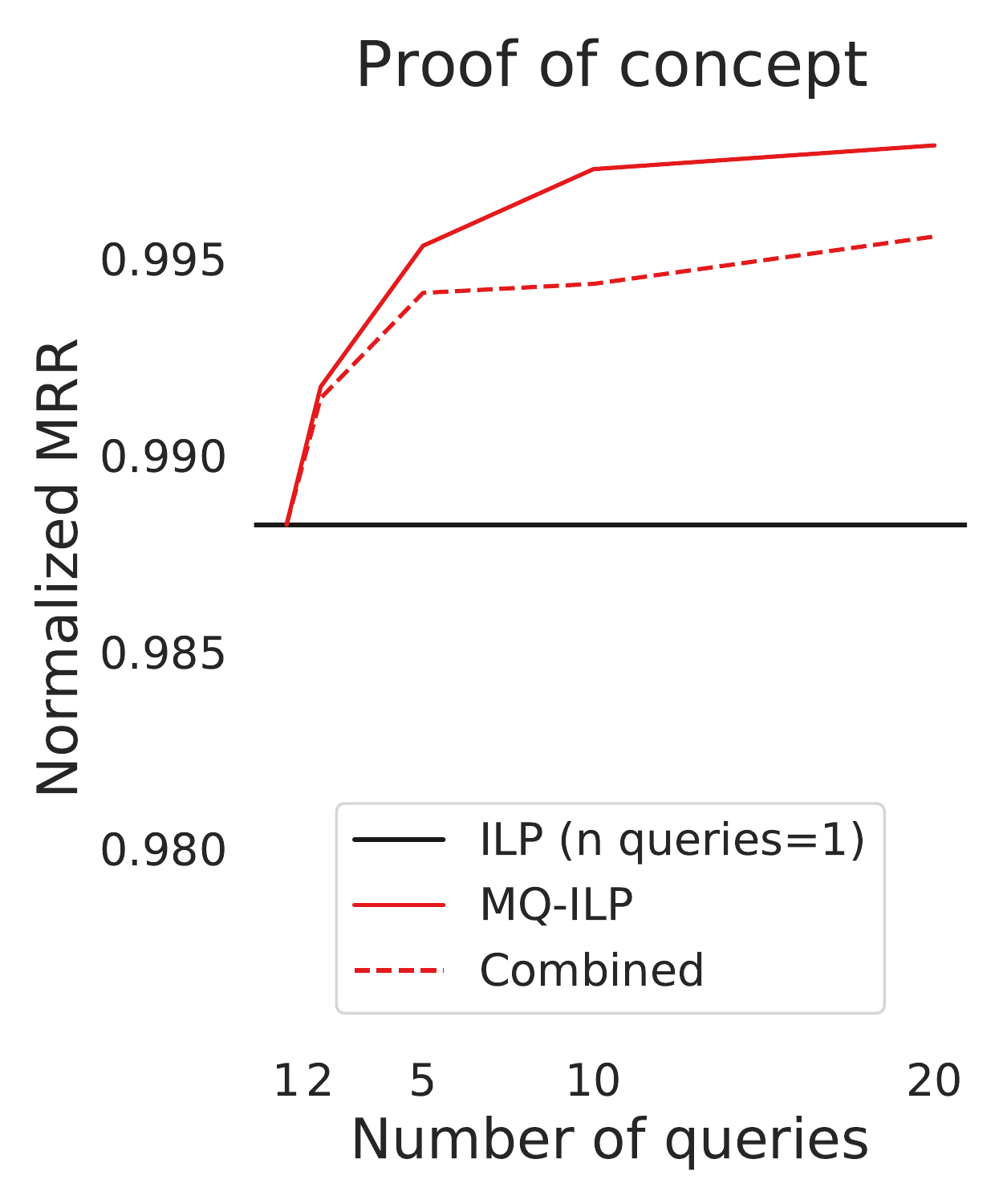}  
          \caption{}
          \label{subfig:no_noise_equal_S}
    \end{subfigure}
    \begin{subfigure}{.33\textwidth}
          \centering
          \includegraphics[width=.9\linewidth]{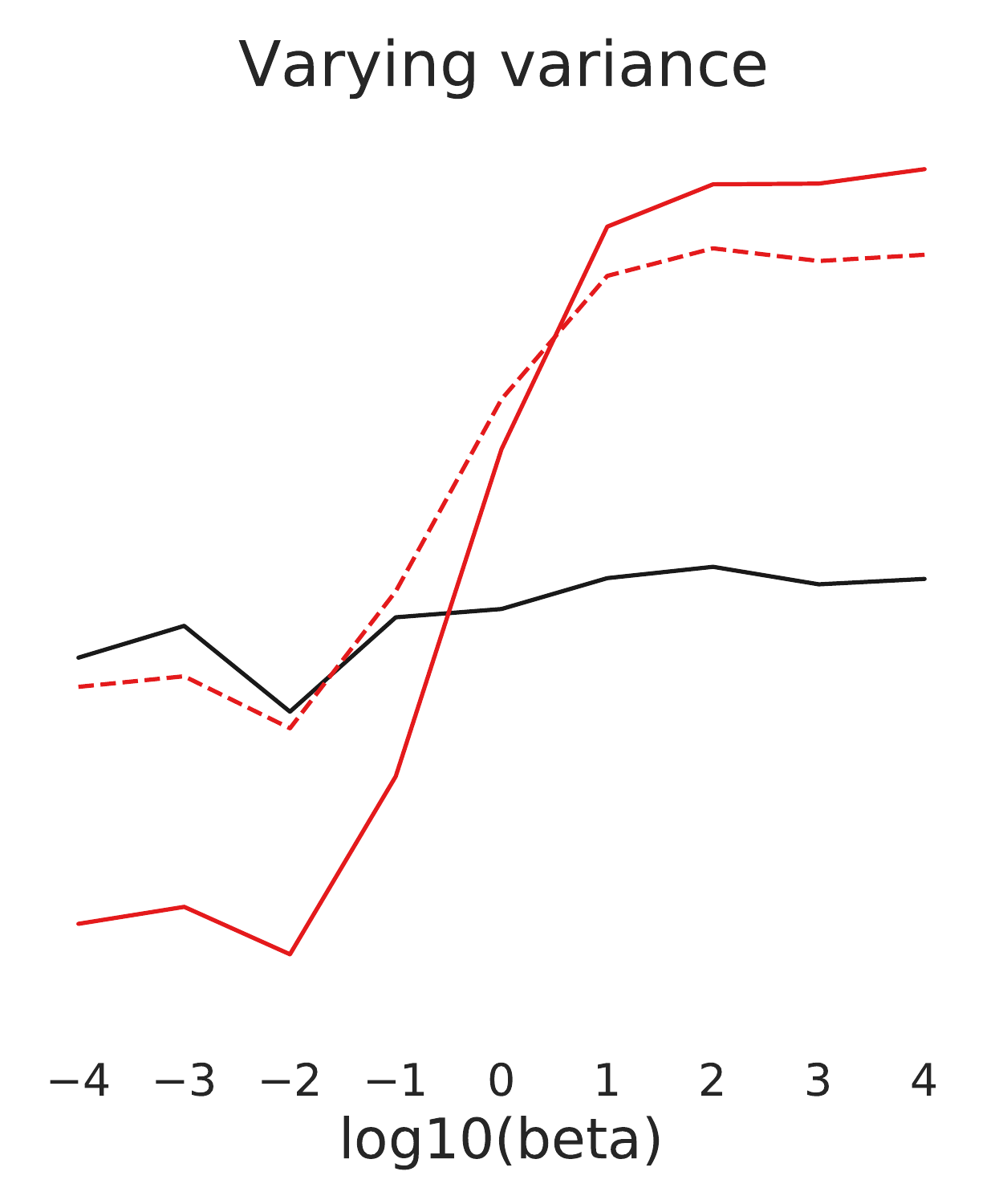}  
          \caption{}
          \label{subfig:varying variance}
    \end{subfigure}
    \begin{subfigure}{.33\textwidth}
          \centering
          \includegraphics[width=.9\linewidth]{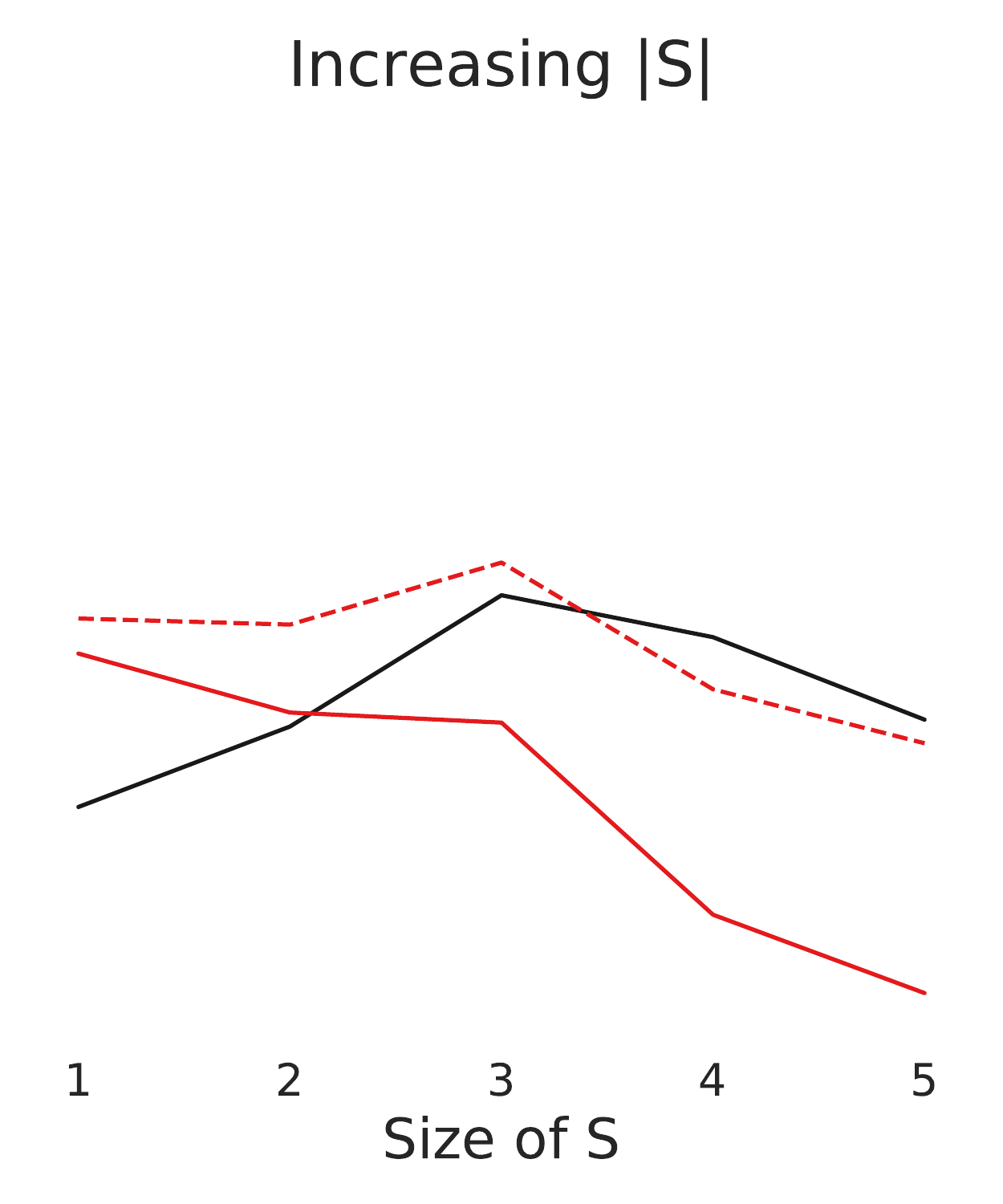}  
          \caption{}
          \label{subfig:inreasing S}
    \end{subfigure}
    \caption{Comparison of i) using only $ (v^{*}, S) $ to learn a distance, ii) using the distance learned from all available queries, iii) using the average of the two learned distances across three simulation settings. The top subfigure outlines a schematic of the simulation: for each query $ q_{k} $ an optimal weight  $\alpha_{k}$ for the dissimilarity corresponding to the adjacency spectral embedding are sampled from a Beta distribution with equal parameters. Subfigure a) is a proof-of-concept and demonstrates that it is possible to gainfully use additional $ (q, S) $ pairs when $ \vec{\alpha} $ is the same for all queries. Subfigure b) shows the effect of variance of the unknown-but-true optimal weight vector by varying $ \beta $ within the $ Beta(\beta, \beta) $ submodel. In particular, when the variance of the unknown-but-true weight vector is large estimate ii) outperforms estimate iii); estimate iii) outperforms ii) when the variance is small. We note that when the variance of $ \vec{\alpha} $ is large the bias of iii) is large and, similarly, when the variance is small the variance of iii) is small. Subfigure c) explicitly demonstrates the "bias-variance" trade-off by increasing the number of elements known to be similar to $ v^{*} $. The decrease in Normalized MRR for all of the algorithms as $ |S| $ increases is an artefact of the decreasing size of the evaluation set as $ |S| $ increases.
    }
\label{fig:fig}
\end{figure*}

\section{An Integer Linear Program: Multiple Queries}

We now present a method to mitigate the two issues mentioned above. In particular, we consider an extension of the setting for which the ILP was proposed -- instead of a single $ (v^{*}, S) $ pair, we now have access to $ K $ (query, set of elements known to be similar to the query) pairs, $ \{(v^{*}, S) \cup \{q_{k}, S_{k}\}_{k=2}^{K} \} $, along with $ J $ dissimilarity measures between the queries and the remaining vertices. It is assumed that the elements of the $ S_{k} $ and $ S $ implicitly define, on average, the same definition of similar for their corresponding query. The goal, as in the setting above, is to produce an effective ranking function relative to $ v^{*} $. For notational convenience we let $ q_{1} := v^{*} $ and $ S_{1} := S $. 

Formally, let $ G = (V, E) $ be a graph 
and 
let $ d^{j}: V \times V \to \mbb{R} $ be the dissimilarity measure corresponding to representation $ j $. As above, we let convex combinations of the $ J $ dissimilarities induce a set of potential ranking functions. We again parameterize these combinations with $ \vec{\alpha} \in \Delta^{J-1} $ and let $ \vec{\alpha}_{1}, \vec{\alpha}_{2}, \hdots, \vec{\alpha}_{K} $ correspond to the combinations associated with $ q_{1}, q_{2}, \hdots, q_{K} $, respectively.

Now suppose that there is a probability distribution over $ (q, S) $ pairs and that $ (q_{k}, S_{k}) \iid P_{(q, S)} $ for some $ P_{(q,S)} $ for $ k = 1, \hdots, K $. We constrain $ P_{(q, S)} $ such that it only puts mass on $ (q, S) $ pairs where the elements of $ S $ are at the top of a ranked list (with respect to $ q $) induced by a convex combination of the available dissimilarities. This restriction allows us to re-parameterize $ P_{(q,S)} $ from a distribution on $ (q, S) $ pairs to a distribution on $ \vec{\alpha} \in \Delta^{J-1} $.

In general, the additional $ (q, S) $ pairs be used to estimate $ P_{\vec{\alpha}} $. In this work we focus only on estimating the the expectation. We let the average true-but-unknown optimal linear combination of the dissimilarities for each vertex $ k $ be $ \bar{\alpha} $. That is, 
\begin{equation}\label{eq:alpha-hat}
    \bar{\alpha} = \mbb{E}_{(q,S) \sim P_{(q, S)}}\left[\argmin_{\vec{\alpha}} \max_{s \in S} h^{\vec{\alpha}}(q, s)\right].
\end{equation}

Since the implicit definitions of similar have the same expectation, ideally a ranking function will map elements of $ S_{k} $ to small values on average across vertices. This leads to the following natural objective function:
\begin{equation}\label{eq:m-ilp}
    \min_{\vec{\alpha}} \sum_{k=1}^{K} \max_{s \in S_{k}} h^{\vec{\alpha}}(q_{k}, s).
\end{equation} As with \eqref{eq:ilp}, the objective function \eqref{eq:m-ilp} can be solved using a mixed integer linear program.  Note that by jointly optimizing as in \eqref{eq:m-ilp} we better regularize than if we were to optimize the ILP objective $ K $ times then average. 

Let $ C_{k} = V \setminus \{\{q_{k}\} \cup S_{k}\} $ be the candidate set for $ q_{k} $. To set up the Multiple Query-ILP (MQ-ILP) we introduce the indicator variables $ x^{\vec{\alpha}}_{v, k} $ for all $ v \in C_{k}$ and all $ q_{k} \in \{q_{1}, \hdots, q_{K}\} $:
\begin{align*}
    x^{\vec{\alpha}}_{v, k} = \begin{cases} 1 \text{ if } h^{\vec{\alpha}}(q_{k}, v) > h^{\vec{\alpha}}(q_{k}, s) \quad \forall s \in S_{k} \\
    0 \text{ o.w.}
    \end{cases}
\end{align*} The MQ-ILP objective function is then
\begin{align*}
    \min_{\vec{\alpha}} \sum_{k=1}^{K} \sum_{v \in C_{k}} x^{\vec{\alpha}}_{v, k}
\end{align*} with the linear constraints
\begin{align*}
\sum_{j=1}^{J} \alpha_{j} d^{j}(q_{k}, s) \leq \sum_{j=1}^{J} \alpha_{j} d^{j}(q_{k}, v) + M\cdot x^{\vec{\alpha}}_{v,k} \\ 
\;\; \forall (s, v) \in (S_{k}, C_{k}) \\
\;\; \forall k \in \{1, \hdots, K\}
\end{align*} to ensure that $ x_{v, k}^{\vec{\alpha}} $ is 0 only when $ v \in C_{k} $ is ranked lower than all elements of $ S_{k} $ and where $ M = \max_{(k,i)} \{\{d^{j}(q_{k}, v_{i})\}_{i=2}^{n}\}_{k=1}^{K} $. Note that the MQ-ILP is equivalent to the ILP when $ K = 1 $ or when $ |S_{k}| = 0 $ for all $ k \in \{2, \hdots, K\} $.

\subsection*{Leveraging the jointly learned distance}
We let the weight vector learned from $ \{(q_{k}, S_{k})\}_{k=1}^{K} $ using the proposed MQ-ILP be $ \hat{\alpha}^{K} $ and the weight vector learned from only $ (q_{1}, S_{1}) = (v^{*}, S) $ be $ \hat{\alpha} $. When ranking elements relative to $ v^{*} $ there are three classes of estimates of the optimal linear combination to consider: i) using only $ \hat{\alpha} $, ii) using only $ \hat{\alpha}^{K} $, and iii) using a combination of $ \hat{\alpha} $ and $ \hat{\alpha}^{K} $.

For i), there are the limitations mentioned above. Namely, the $ S $ set is sometimes too noisy to learn an effective ranking scheme relative to $ v^{*} $.

For ii), the effectiveness of the induced ranking function is a function of the distributional properties of $ P_{\alpha} $. In particular, as we demonstrate in simulations below, the ranking scheme induced by $ \hat{\alpha}^{K} $ is effective in cases where $ \alpha \sim P_{\alpha} $ has a small variance. Otherwise, the bias in the estimate $ \hat{\alpha} $ relative to the true-but-unknown optimal weight vector for $ v^{*} $ renders the jointly learned weight vector ineffective.

For iii), with no additional assumptions made on the properties of $ P_{\alpha} $ there is no optimal way to combine $ \hat{\alpha} $ and $ \hat{\alpha}^{K} $. While we do not pursue adding structure to $ P_{\alpha} $ we note, and demonstrate in simulations, that a simple average of the two estimates yields a combination that can be effective. Investigating more elaborate combinations of the two estimates as a function of the data is a promising line of future research.

From the perspective of estimating the true-but-unknown optimal weight vector $ \vec{\alpha} $ for $ v^{*} $, we (loosely) think of the three classes of estimates i), ii), and iii) as having different qualities with respect to the ``bias-variance tradeoff" \citep{hastie2009elements}:
\renewcommand{\theenumi}{\roman{enumi}}
\begin{enumerate}
    \item has low bias and high variance since it is only using data directly relevant to estimating $ \vec{\alpha} $ (low bias) but has access to a small amount of data (high variance);
    \item has high bias and low variance since it is using data not directly relevant to estimating $ \vec{\alpha} $ (high bias) but has access to more data (low variance);
    \item has ``medium" bias and ``medium" variance since it can interpolate between i) and ii). 
\end{enumerate}
\begin{figure*}
    \captionsetup[subfigure]{justification=centering}     
    \begin{subfigure}{\textwidth}
          \centering
        \includegraphics[width=\linewidth]{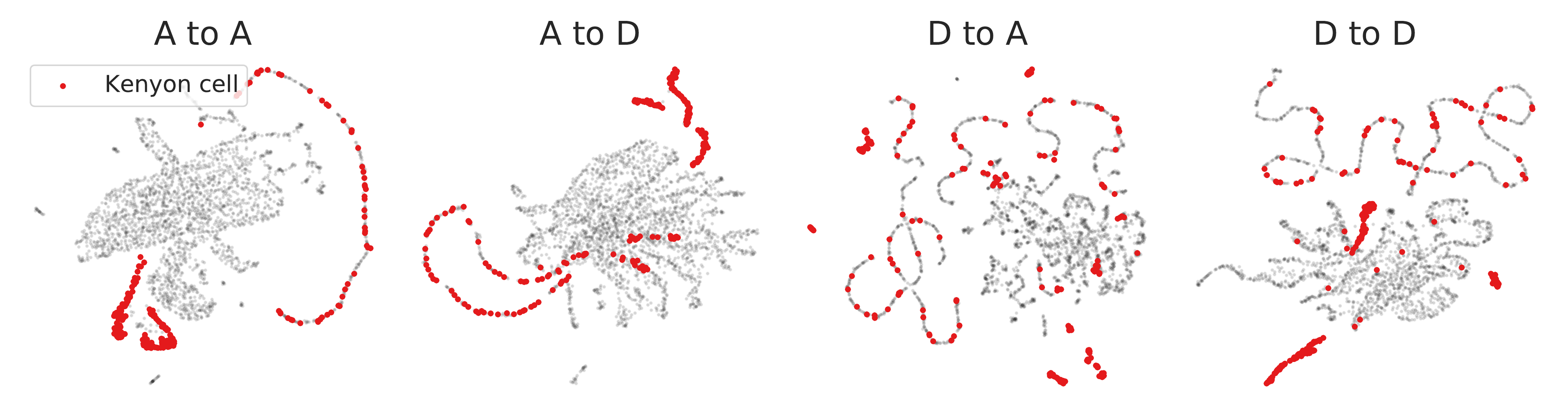}  
          \caption{}
    \end{subfigure} \\
    \centering
    \begin{subfigure}{0.49\textwidth}
          \centering
          \includegraphics[width=\linewidth]{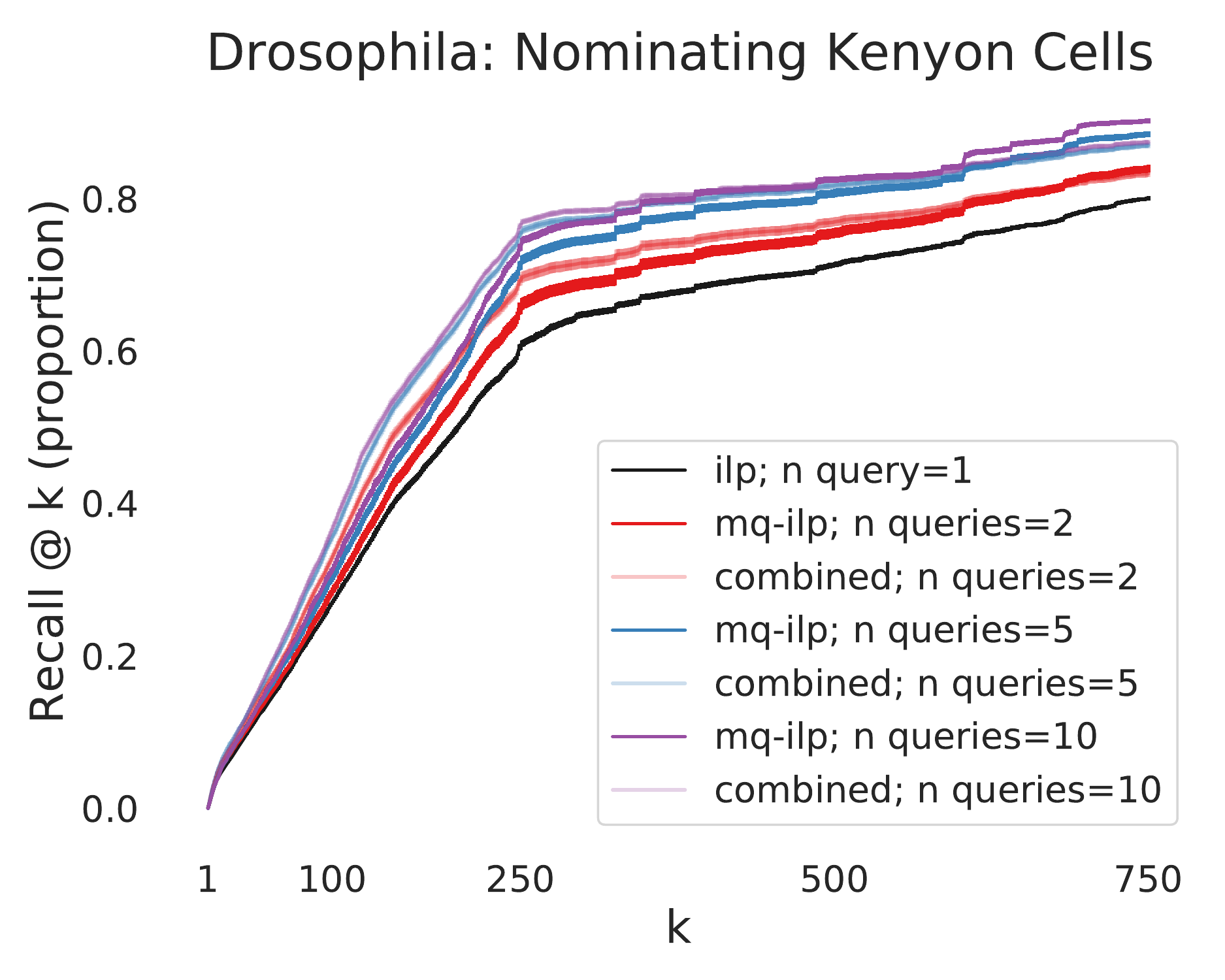}  
          \caption{}
          \label{subfig:drosophila-poc}
    \end{subfigure}
    \caption{Applying the MQ-ILP to ranking Kenyon cells in the mushroom body of the \textit{Drosophila} larva connectome. The top row of figures shows two-dimensional representations of the four different embeddings (from unique edge types) used to learn the distance for which to deploy for ranking. The two-dimensional representations were learned using UMAP. The figure shows that it is possible to use the distance learned from the MQ-ILP to improve ranking performance as measured by recall at $ k $. The lines are the average recall at $ k $ estimated using 1000 Monte Carlo iterations. Error bars represent the standard error of the estimated means. 
    }
\label{fig:drosophila}
\end{figure*}

\section{Simulations using the Random Dot Product Graph} The set of simulation settings we consider is designed to explore the trade-offs of the ranking functions induced by i), ii), and iii) as a function of relevant hyperparameters: the number of queries $ K $, the size of the $ S $, the size of $ S_{k} $, and the distribution of $ \alpha $. 

To facilitate this we use a latent space network model \citep{hoff2002latent}. Recall that in the latent space network model the probability of the existence of an edge is based on a kernel function between two vectors, often called \textit{latent positions}, one representing each of the two vertices. We denote the matrix containing all of the pairwise edge probabilities with $ P $. Given $ P $, edges are sampled independently $ Bernoulli(P_{ij}) $. 

One such latent space network model is the Random Dot Product Graph (RDPG) model \citep{athreya2017statistical}. In the RDPG model the kernel is simply the dot product between two vectors in $ \mbb{R}^{m} $. Hence, with $ X_{i}, X_{j} \in \mbb{R}^{m} $ and $ 0 \le \langle X_{i}, X_{j}\rangle \le 1 $ for all $ (i, j) $ the RDPG is such that $ P_{ij} = \langle X_{i}, X_{j}\rangle $. 

In all of the simulations below, we first generate $ n = 200 $ latent positions $ i.i.d. $ according to the uniform distribution on the positive unit disk in $ \mbb{R}^{2} $. We then consider $ J = 2 $ transformations of $ P $, the Adjacency Spectral Embedding (ASE) \citep{doi:10.1080/01621459.2012.699795} and the Laplacian Spectral Embedding (LSE) \citep{von2007tutorial}. Both ASE and LSE can yield vector representations of the vertices in $ \mbb{R}^{2} $. With the vector representations of the vertices we then calculate the pairwise (Euclidean) distances and use them, as required, as input to the ILP and MQ-ILP. We let $ d_{\{ASE, LSE\}} $ denote the Euclidean distance between two vertices in the vector representations of \{ASE, LSE\}. We chose ASE and LSE as the base representations of the vertices because of their tractable analytical properties under the RDPG model \citep{doi:10.1080/01621459.2012.699795, athreya2017statistical}. 

Once $ P $ is sampled, $ K $ different $ (q, S) $ pairs are randomly selected. One of the $ K $ is then randomly selected as $ (v^{*}, S) $. In particular, the query $ q $ is randomly selected without replacement from the set of $ 200 $ vertices. Then, for each particular query, a weight vector $ \alpha $ is sampled $ i.i.d $ from $ P_{\alpha} $. If $ q = v^{*} $ then $ |S| $ of the closest 10 vertices to $ q $, according to the dissimilarity $ d_{\alpha} = \alpha d_{ASE} + (1-\alpha) d_{LSE} $, are chosen as elements of $ S $. Otherwise, $ |S_{k}| $ of the closest 10 vertices to $ q $, according to $ d_{\alpha} $, are chosen as elements of $ S_{k} $. 

We evaluate three different methods via normalized MRR on the 5 closest elements to $ v^{*} $ according to $ (q, S) $ that are not elements of $ S $. The three methods are i) the ranking function induced by $ \hat{\alpha} $, ii) the ranking function induced by $ \hat{\alpha}^{K} $, and iii) the ranking function induced by $ \frac{1}{2}\left( \hat{\alpha} + \hat{\alpha}^{K}\right) $.

We implemented these simulations with the aid of the network and graph statistics Python package \texttt{graspologic} \citep{chung2019graspy}.

\subsection{A proof of concept}
We begin by letting $ P_{\alpha} $ be a point mass at $ \alpha = 0.5 $ and setting $ |S| = |S_{k}| = 3 $. Since $ |S| = |S_{k}| $ and all of the definitions of similar are the same, using the estimate of $ \alpha $ that uses more data ($ \hat{\alpha}^{K} $) dominates both the estimate from using only $ (v^{*}, S) $ and the combined estimate from the perspective or normalized mean reciprocal rank, as seen in Figure \ref{subfig:no_noise_equal_S}. 

\subsection{Varying variance}
Next, we let $ P_{\alpha} = Beta(\beta, \beta) $ with $ \beta \in \{10^{i}\}_{i=-4}^{4} $. We continue to let $ |S| = |S_{k}| = 3 $ and fix $ K = 10 $. We note that as $ \beta $ goes from $ 10^{-4} $ to $ 10^{4} $ the expected value of $ \alpha $ stays constant ($\mbb{E}(\alpha) = 0.5$) while the variance goes from close to maximal $ (\approx \frac{1}{4} $ when $ \beta = 10^{-4}) $ to minimal $ (\approx 0 $ when $ \beta = 10^{4}) $. The effectiveness of the additional queries increases as the variance of $ \alpha $ decreases, as seen in Figure \ref{subfig:varying variance}. We note the regime change near $ \beta = 1 $ where the estimate from the MQ-ILP begins to outperform the combined estimate. Loosely speaking, this occurs once the bias of the MQ-ILP is small relative to the variance of the estimate from the ILP. 

\subsection{Increasing the size of $ S $}
Next, we fix $ P_{\alpha} = Beta(10^{-1}, 10^{-1}) $, $ K = 10 $, $ |S_{k}| = 3 $ and vary $ |S| \in \{1, \hdots, 10\} $. As $ |S| $ increases there are three noteworthy regimes, as seen in Figure \ref{subfig:inreasing S}. First, when $ |S| = 1 $ and the variance of the estimate from using only $ (v^{*}, S) $ is high, the combined estimate outperforms the MQ-ILP estimate which outperforms the ILP estimate. Second, when $ |S| \in \{2, 3, 4\} $ and the variance decreases a little, the ILP estimate outperforms the MQ-ILP estimate. Third, once $ |S| > 4 $, the ILP outperforms both estimates because it is both low variance and low bias.


\begin{figure*}
    \captionsetup[subfigure]{justification=centering}     
    \begin{subfigure}{\textwidth}
          \centering
         \includegraphics[width=\linewidth]{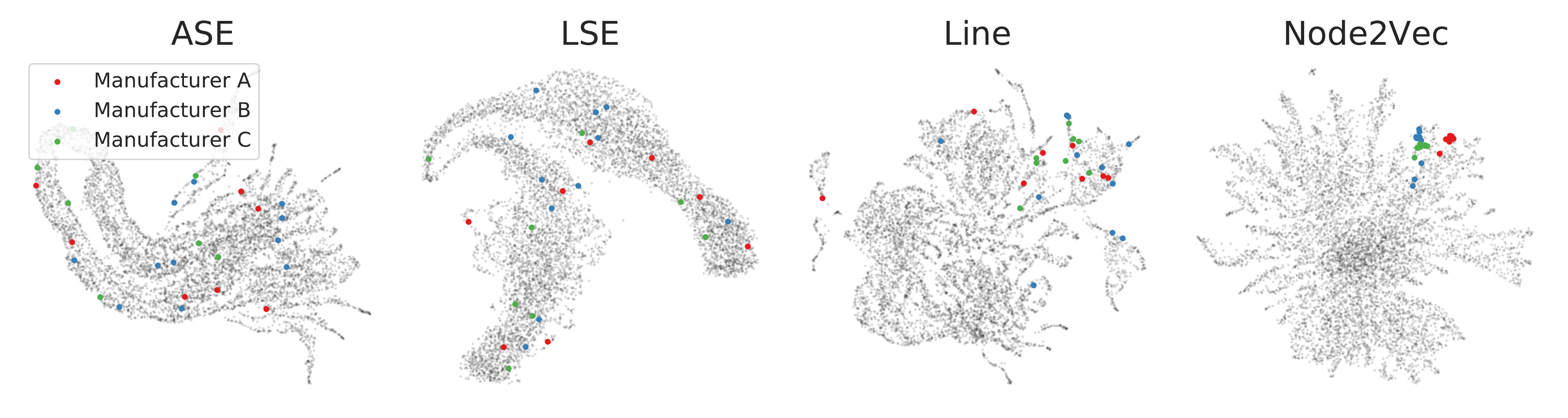}  
          \caption{}
    \end{subfigure} \\
    \centering
    \begin{subfigure}{.5\textwidth}
          \centering
          \includegraphics[width=\linewidth]{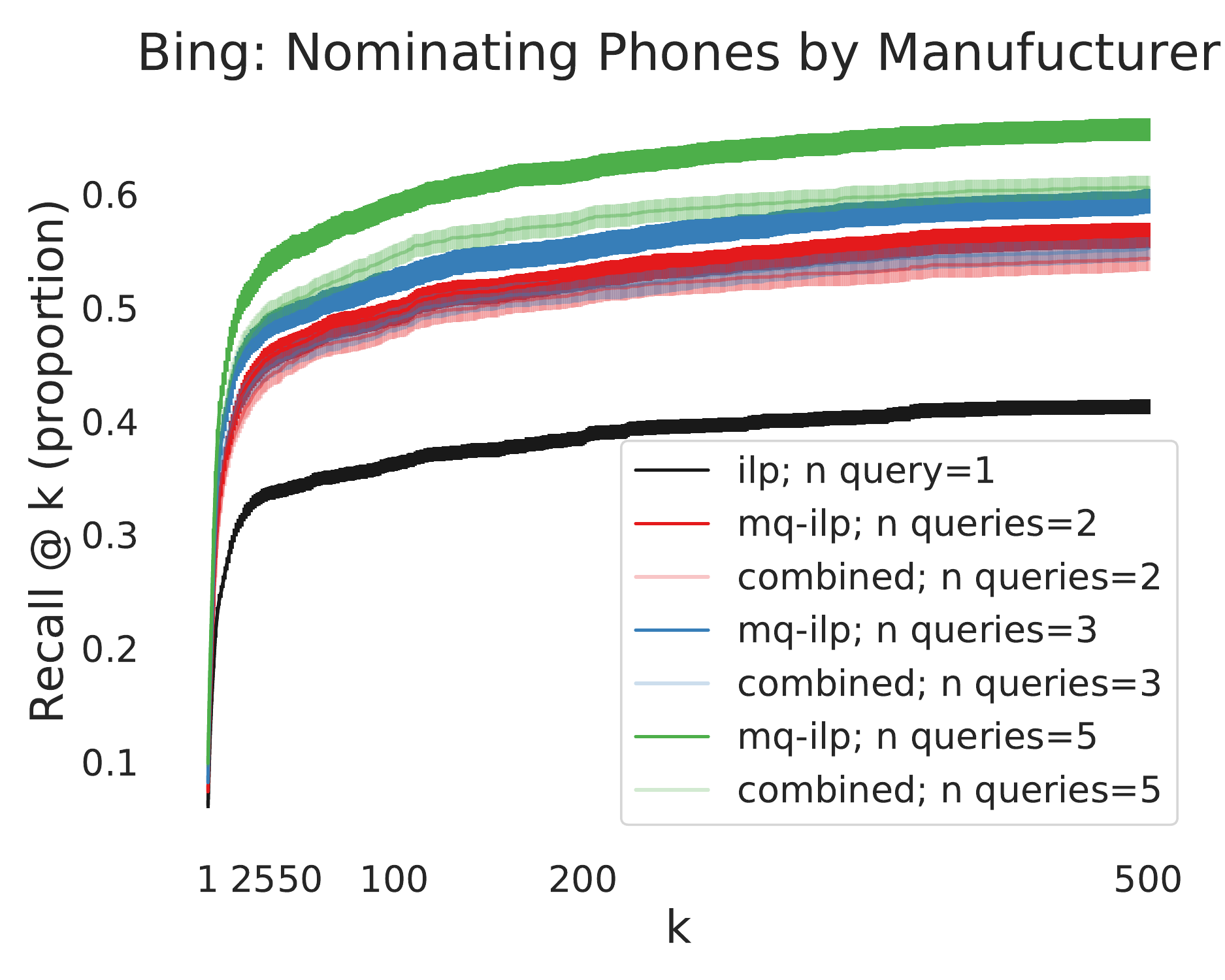}  
          \caption{}
          \label{subfig:bing-poc}
    \end{subfigure}
    \caption{Applying the MQ-ILP to ranking phones in a Bing navigational graph. The top row of the figures shows two-dimensional representations of the four different embeddings used to learn the distance for which to deploy for ranking. The bottom row again demonstrates the leveraging the semantically similar nodes, both via the MQ-ILP and the combined estimates, improves performance over the ILP. The lines are the average recall at $ k $ estimated using 1000 Monte Carlo iterations. Error bars represent the standard error of the estimated means.}
\label{fig:bing}
\end{figure*}

\section{The \textit{Drosophila} larvae connectome and a Bing navigational graph}

We consider two real data examples that demonstrate the effectiveness of the proposed MQ-ILP: one using part of a high-resolution connectome of a \textit{Drosophila} larvae and one using a search navigational graph from Bing.

We compare and evaluate $ \hat{\alpha} $, $ \hat{\alpha}^{K} $ and $ \frac{1}{2}(\hat{\alpha} + \hat{\alpha}^{K}) $ using Recall at $k$. As far as we are aware, there are no methods in the literature that explicitly handle dissimilarity-based learning for ranking from only positive examples. Other natural competitors, such as randomly selecting weight vectors and decomposing the weight matrix, are explored in \citep{Helm2020LearningTR}.


\subsection*{\textit{Drosophila}}
We first apply the MQ-ILP to the task of nominating Kenyon cells (KC) in the mushroom body \citep{eichler2017complete} of the \textit{Drosophila} larva synaptic-level connectome. A synaptic-level connectome \citep{vogelstein2019connectal} is typically a network defined on a set of neurons where a (directed and weighted) edge exists between neuron $ i $ and neuron $ j $ if neuron $ i $ synapses onto neuron $ j $. 

However, a neuron is often thought of as two entities, a dendrite and an axon, and it is possible to measure which axon synapses onto which axon, which axon synapses onto which dendrite, and so on. This leads to $ 4 $ different synapse-level connectomes (data unpublished): $ G_{AA} = (V, E_{AA}), G_{AD}, G_{DA}, G_{DA} $. In each of the networks there are $ n = 2965 $ entities and the individual edge counts are $|E_{AD}| = 54303,  |E_{AA}| = 34867, |E_{DD}| = 10209$, and $ |E_{DA}| = 4088 $. These networks were manually annotated from electron microscopy imagery for a single \textit{Drosophila} larva brain \citep{ohyama2015multilevel, schneider2016quantitative}.

For each of the synaptic-level connectomes, we consider the pairwise Euclidean distances of the spectral embeddings of the normalized Laplacians ($\text{embedding dimension} = 11 $), leading to $ J = 4 $ different representations of the $ 2965 $ vertices. Two-dimensional representations, learned using UMAP \citep{mcinnes2018umap} as implemented in \citep{pedregosa2011scikit} in Python, of the $11$-dimensional representations used are shown in the top row of Figure \ref{fig:drosophila}. Notably, the different networks yield non-trivially different embeddings and hence possibly different rankings of the pairwise distances. 


We investigate the effect of increasing the number of semantically similar queries. To do this we first randomly sample a $ v^{*} $. The $ v^{*} $ is either in the left or right hemisphere of the connectome. We then sample (without replacement) $ K \in \{2,5,10\} $ different Kenyon cells from the same hemisphere as $ v^{*} $. 
For $ v^{*} $ we randomly sample (without replacement) $ |S| = 5 $ other KCs as elements of $ S $. For each of the remaining sampled $ K $ KCs we randomly sample (without replacement) $ |S_{k}| = 5 $ other KCs as elements of $ S_{k} $. The ranking functions induced by the three estimates of the optimal combination are evaluated via Recall at $ k $ for various $ k $ on $ 15 $ (different for each iteration) KCs. 

Figure \ref{fig:drosophila} shows that using more semantically similar queries to learn a distance improves performance when nominating Kenyon cells. Further, for small $ k $ the combined estimates unilaterally outperform the MQ-ILP estimates as measured by recall at $ k $. This could mean that the optimal weight vectors for the semantically similar queries are not identical to the optimal weight vector for $ v^{*} $ and that weighing the estimate from the ILP and the estimate from the MQ-ILP equally both reduces bias relative to the estimate from the MQ-ILP and reduces variance relative to the estimate from the ILP. 




\subsection*{Bing}

We next apply the MQ-ILP to the task of nominating phones within $ G_{Bing} $, a network of devices. $ G_{Bing} $ is derived from navigational data from Bing. In particular, each vertex in the network is a searchable object, i.e. ``iPhone X", and a (directed) edge $ (i, j) $ exists from object $ i $ to object $ j $ if a user searches for object $ i $ and subseqeuently searches for object $ j $ within the same browsing session. Edges are weighted by the frequency of the occurrence of the subsequent search. This raw network is pre-processed by a team within Microsoft. The network we study, $ G_{Bing} $, is then the largest connected component of the subgraph containing the devices (phones, tablets, laptops, etc.). $ G_{Bing} $ has $ |V| = 7,876 $ vertices and $ |E| = 46,062 $ directed and weighted edges. 

We are given $ J = 4 $ different pairwise distances of the vertices of $ G_{Bing} $ derived from four different representations: ASE \citep{doi:10.1080/01621459.2012.699795} ($ \text{embedding dimension} = 32 $), LSE \citep{von2007tutorial} ($ \text{embedding dimension} = 32 $), Node2Vec \citep{grover2016node2vec} ($ \text{embedding dimension} = 32 $), and Verse \citep{tsitsulin2018verse} ($ \text{embedding dimension} = 64 $).

The top row of Figure \ref{fig:bing} shows a two-dimensional representation of the $ J=4$ different representations. The two-dimensional representation is a projection learned using tSNE. 

Of the $ 7,876 $ vertices we are given $ 27 $ phones -- $ 8 $ from manufacturer $ A $, $ 8 $ from manufacturer $ B $, and $ 10 $ from manufacturer $ C $ -- for which we are tasked with creating rank lists of the remaining objects, ideally with objects most similar to the phone concentrating near the top of the rank list. We again emulate the first (proof of concept; Figure \ref{subfig:no_noise_equal_S}) and third (increasing $|S|$; Figure \ref{subfig:inreasing S}) simulation settings.

For the first setting we investigate the effect of increasing $ K $, the number of (query, S) pairs. In particular, first we randomly select a phone. Then, we randomly select $ |S| = 2 $ phones from the same manufacturer for each of the $ K $ phones. For example, for $ v^{*} = $``iPhone X" its corresponding $ S $ set may be \{``iPhone Xs", ``iPhone 11"\}. We then use this $ (v^{*}, S) $ pair as input to the ILP to learn a ranking scheme relative to $ v^{*}$. For input to the MQ-ILP we sample $ K-1 \in \{1, \hdots, 6\} $ phones from the same manufacturer and for each the sampled phones subsequently sample $ |S_{k}| = 2 $ phones from the same manufacturer without replacement. We evaluate the three methods via Recall at $ k $ for various $ k $ on a three phones from the same manufacturer as $ v^{*} $. 

As we see in the simulations setting, we see in ranking Kenyon cells in the mushroom body of the \textit{Drosophila} connectome and phones in the navigational graph -- leveraging semantically similar vertices improves performance. Indeed, in the real data settings we studied the more semantically similar vertices the MQ-ILP has access to the better it performs.

\section{Discussion and conclusion}
We proposed and evaluated a novel ranking algorithm, the MQ-ILP, that finds an optimal combination of representations across a set of semantically similar queries. 

In simulation we showed that the MQ-ILP is an effective alternative to the previously studied ILP across a wide variety of regimes. Further, the MQ-ILP improves upon previous results for ranking Kenyon cells in the mushroom body of the connectome of the \textit{Drosophila} larvae and for ranking phones in the Bing navigational graph. The combination of these results demonstrate the utility of the MQ-ILP across a range of applications.

For example, the MQ-ILP is a natural candidate for ranking in privacy-constrained applications -- the ranking functions learned by the MQ-ILP and ILP are weighted averages of existing representations; adding zero-centered noise to each representation would affect the learned ranking functions less than, say, using the optimal single representation. Thus the ILP and MQ-ILP are likely well-positioned on the privacy-performance curve and the application of the MQ-ILP in those settings is an interesting and promising line of future research.

Further extensions of this work include adapting the MQ-ILP to active or online settings where at each step a set of previously unlabeled examples are chosen to be labeled either positive or negative. If the negatively labeled examples are ignored, this extension could be as simple as using the previous optimal solution as an initialization for the current iteration. Somehow including the negative information in the optimization, however, would likely improve ranking performance. Hence, exploring different mechanisms to ``push negative examples down" the ranked list -- similar to how the MQ-ILP ``pushes positive examples up" the ranked list could be a fruitful direction for active learning.

Lastly, while we do not pursue theoretical results of the MQ-ILP here, we think results that comment on the probability of a unknown-to-be-similar element of the nomination set being in the top $ k $ of the ranked list induced by the optimal combination of representations are possible to obtain in particular settings. The simulation settings herein provide a good place to start for theoretical directions because of the analytical properties of ASE and LSE under the RDPG.

\ifCLASSOPTIONcompsoc
  \section*{Acknowledgments}
\else
  \section*{Acknowledgment}
\fi

The authors would like to thank Anton Alyakin for his comments on an earlier draft of this manuscript.

\ifCLASSOPTIONcaptionsoff
  \newpage
\fi

\clearpage



\bibliographystyle{IEEEtran}
\bibliography{biblio.bib}

\begin{thebibliography}{10}
\providecommand{\url}[1]{#1}
\csname url@samestyle\endcsname
\providecommand{\newblock}{\relax}
\providecommand{\bibinfo}[2]{#2}
\providecommand{\BIBentrySTDinterwordspacing}{\spaceskip=0pt\relax}
\providecommand{\BIBentryALTinterwordstretchfactor}{4}
\providecommand{\BIBentryALTinterwordspacing}{\spaceskip=\fontdimen2\font plus
\BIBentryALTinterwordstretchfactor\fontdimen3\font minus
  \fontdimen4\font\relax}
\providecommand{\BIBforeignlanguage}[2]{{%
\expandafter\ifx\csname l@#1\endcsname\relax
\typeout{** WARNING: IEEEtran.bst: No hyphenation pattern has been}%
\typeout{** loaded for the language `#1'. Using the pattern for}%
\typeout{** the default language instead.}%
\else
\language=\csname l@#1\endcsname
\fi
#2}}
\providecommand{\BIBdecl}{\relax}
\BIBdecl

\bibitem{10.1007/s10791-009-9123-y}
\BIBentryALTinterwordspacing
T.~Qin, T.-Y. Liu, J.~Xu, and H.~Li, ``Letor: A benchmark collection for
  research on learning to rank for information retrieval,'' \emph{Inf. Retr.},
  vol.~13, no.~4, p. 346–374, Aug. 2010. [Online]. Available:
  \url{https://doi.org/10.1007/s10791-009-9123-y}
\BIBentrySTDinterwordspacing

\bibitem{pmlr-v14-chapelle11a}
\BIBentryALTinterwordspacing
O.~Chapelle and Y.~Chang, ``Yahoo! learning to rank challenge overview,'' ser.
  Proceedings of Machine Learning Research, O.~Chapelle, Y.~Chang, and T.-Y.
  Liu, Eds., vol.~14.\hskip 1em plus 0.5em minus 0.4em\relax Haifa, Israel:
  PMLR, 25 Jun 2011, pp. 1--24. [Online]. Available:
  \url{http://proceedings.mlr.press/v14/chapelle11a.html}
\BIBentrySTDinterwordspacing

\bibitem{fishkind2015vertex}
D.~E. Fishkind, V.~Lyzinski, H.~Pao, L.~Chen, C.~E. Priebe \emph{et~al.},
  ``Vertex nomination schemes for membership prediction,'' \emph{The Annals of
  Applied Statistics}, vol.~9, no.~3, pp. 1510--1532, 2015.

\bibitem{pan2009survey}
S.~J. Pan and Q.~Yang, ``A survey on transfer learning,'' \emph{IEEE
  Transactions on knowledge and data engineering}, vol.~22, no.~10, pp.
  1345--1359, 2009.

\bibitem{wang2018deep}
M.~Wang and W.~Deng, ``Deep visual domain adaptation: A survey,''
  \emph{Neurocomputing}, vol. 312, pp. 135--153, 2018.

\bibitem{vogelstein2020general}
J.~T. Vogelstein, H.~S. Helm, R.~D. Mehta, J.~Dey, W.~LeVine, W.~Yang,
  B.~Tower, J.~Larson, C.~White, and C.~E. Priebe, ``A general approach to
  progressive learning,'' 2020.

\bibitem{Helm2020LearningTR}
H.~S. Helm, A.~Basu, A.~Athreya, Y.~Park, J.~Vogelstein, M.~Winding, M.~Zlatic,
  A.~Cardona, P.~Bourke, J.~Larson, C.~White, and C.~Priebe, ``Learning to rank
  via combining representations,'' \emph{ArXiv}, vol. abs/2005.10700, 2020.

\bibitem{JMLR:v20:18-048}
\BIBentryALTinterwordspacing
V.~Lyzinski, K.~Levin, and C.~E. Priebe, ``On consistent vertex nomination
  schemes,'' \emph{Journal of Machine Learning Research}, vol.~20, no.~69, pp.
  1--39, 2019. [Online]. Available:
  \url{http://jmlr.org/papers/v20/18-048.html}
\BIBentrySTDinterwordspacing

\bibitem{pmlr-v25-zhou12}
\BIBentryALTinterwordspacing
J.~T. Zhou, S.~J. Pan, Q.~Mao, and I.~W. Tsang, ``Multi-view positive and
  unlabeled learning,'' ser. Proceedings of Machine Learning Research, S.~C.~H.
  Hoi and W.~Buntine, Eds., vol.~25.\hskip 1em plus 0.5em minus 0.4em\relax
  Singapore Management University, Singapore: PMLR, 04--06 Nov 2012, pp.
  555--570. [Online]. Available:
  \url{http://proceedings.mlr.press/v25/zhou12.html}
\BIBentrySTDinterwordspacing

\bibitem{liu2011learning}
T.-Y. Liu, \emph{Learning to rank for information retrieval}.\hskip 1em plus
  0.5em minus 0.4em\relax Springer Science \& Business Media, 2011.

\bibitem{conte2004thirty}
D.~Conte, P.~Foggia, C.~Sansone, and M.~Vento, ``Thirty years of graph matching
  in pattern recognition,'' \emph{International journal of pattern recognition
  and artificial intelligence}, vol.~18, no.~03, pp. 265--298, 2004.

\bibitem{Bekker_2020}
\BIBentryALTinterwordspacing
J.~Bekker and J.~Davis, ``Learning from positive and unlabeled data: a
  survey,'' \emph{Machine Learning}, vol. 109, no.~4, p. 719–760, Apr 2020.
  [Online]. Available: \url{http://dx.doi.org/10.1007/s10994-020-05877-5}
\BIBentrySTDinterwordspacing

\bibitem{10.1145/1401890.1401920}
\BIBentryALTinterwordspacing
C.~Elkan and K.~Noto, ``Learning classifiers from only positive and unlabeled
  data,'' in \emph{Proceedings of the 14th ACM SIGKDD International Conference
  on Knowledge Discovery and Data Mining}, ser. KDD ’08.\hskip 1em plus 0.5em
  minus 0.4em\relax New York, NY, USA: Association for Computing Machinery,
  2008, p. 213–220. [Online]. Available:
  \url{https://doi.org/10.1145/1401890.1401920}
\BIBentrySTDinterwordspacing

\bibitem{radev2002evaluating}
D.~R. Radev, H.~Qi, H.~Wu, and W.~Fan, ``Evaluating web-based question
  answering systems.'' in \emph{LREC}, 2002.

\bibitem{coppersmith2014vertex}
G.~Coppersmith, ``Vertex nomination,'' \emph{Wiley Interdisciplinary Reviews:
  Computational Statistics}, vol.~6, no.~2, pp. 144--153, 2014.

\bibitem{eichler2017complete}
K.~Eichler, F.~Li, A.~Litwin-Kumar, Y.~Park, I.~Andrade, C.~M.
  Schneider-Mizell, T.~Saumweber, A.~Huser, C.~Eschbach, B.~Gerber
  \emph{et~al.}, ``The complete connectome of a learning and memory centre in
  an insect brain,'' \emph{Nature}, vol. 548, no. 7666, pp. 175--182, 2017.

\bibitem{marchette2011vertex}
D.~Marchette, C.~Priebe, and G.~Coppersmith, ``Vertex nomination via attributed
  random dot product graphs,'' in \emph{Proceedings of the 57th ISI World
  Statistics Congress}, vol.~6, 2011, p.~16.

\bibitem{coppersmith2012vertex}
G.~A. Coppersmith and C.~E. Priebe, ``Vertex nomination via content and
  context,'' \emph{arXiv preprint arXiv:1201.4118}, 2012.

\bibitem{sun2012comparison}
M.~Sun, M.~Tang, and C.~E. Priebe, ``A comparison of graph embedding methods
  for vertex nomination,'' in \emph{2012 11th International Conference on
  Machine Learning and Applications}, vol.~1.\hskip 1em plus 0.5em minus
  0.4em\relax IEEE, 2012, pp. 398--403.

\bibitem{suwan2015bayesian}
S.~Suwan, D.~S. Lee, and C.~E. Priebe, ``Bayesian vertex nomination using
  content and context,'' \emph{Wiley Interdisciplinary Reviews: Computational
  Statistics}, vol.~7, no.~6, pp. 400--416, 2015.

\bibitem{agterberg2019vertex}
J.~Agterberg, Y.~Park, J.~Larson, C.~White, C.~E. Priebe, and V.~Lyzinski,
  ``Vertex nomination, consistent estimation, and adversarial modification,''
  2019.

\bibitem{yoder2020vertex}
J.~Yoder, L.~Chen, H.~Pao, E.~Bridgeford, K.~Levin, D.~E. Fishkind, C.~Priebe,
  and V.~Lyzinski, ``Vertex nomination: The canonical sampling and the extended
  spectral nomination schemes,'' \emph{Computational Statistics \& Data
  Analysis}, vol. 145, p. 106916, 2020.

\bibitem{von2007tutorial}
U.~Von~Luxburg, ``A tutorial on spectral clustering,'' \emph{Statistics and
  computing}, vol.~17, no.~4, pp. 395--416, 2007.

\bibitem{grover2016node2vec}
A.~Grover and J.~Leskovec, ``node2vec: Scalable feature learning for
  networks,'' in \emph{Proceedings of the 22nd ACM SIGKDD international
  conference on Knowledge discovery and data mining}, 2016, pp. 855--864.

\bibitem{tsitsulin2018verse}
A.~Tsitsulin, D.~Mottin, P.~Karras, and E.~M{\"u}ller, ``Verse: Versatile graph
  embeddings from similarity measures,'' in \emph{Proceedings of the 2018 World
  Wide Web Conference}, 2018, pp. 539--548.

\bibitem{hamilton2017inductive}
W.~Hamilton, Z.~Ying, and J.~Leskovec, ``Inductive representation learning on
  large graphs,'' in \emph{Advances in neural information processing systems},
  2017, pp. 1024--1034.

\bibitem{doi:10.1080/01621459.2012.699795}
\BIBentryALTinterwordspacing
D.~L. Sussman, M.~Tang, D.~E. Fishkind, and C.~E. Priebe, ``A consistent
  adjacency spectral embedding for stochastic blockmodel graphs,''
  \emph{Journal of the American Statistical Association}, vol. 107, no. 499,
  pp. 1119--1128, 2012. [Online]. Available:
  \url{https://doi.org/10.1080/01621459.2012.699795}
\BIBentrySTDinterwordspacing

\bibitem{qiu2019netsmf}
J.~Qiu, Y.~Dong, H.~Ma, J.~Li, C.~Wang, K.~Wang, and J.~Tang, ``Netsmf:
  Large-scale network embedding as sparse matrix factorization,'' in \emph{The
  World Wide Web Conference}, 2019, pp. 1509--1520.

\bibitem{hastie2009elements}
T.~Hastie, R.~Tibshirani, and J.~Friedman, \emph{The elements of statistical
  learning: data mining, inference, and prediction}.\hskip 1em plus 0.5em minus
  0.4em\relax Springer Science \& Business Media, 2009.

\bibitem{hoff2002latent}
P.~D. Hoff, A.~E. Raftery, and M.~S. Handcock, ``Latent space approaches to
  social network analysis,'' \emph{Journal of the american Statistical
  association}, vol.~97, no. 460, pp. 1090--1098, 2002.

\bibitem{athreya2017statistical}
A.~Athreya, D.~E. Fishkind, M.~Tang, C.~E. Priebe, Y.~Park, J.~T. Vogelstein,
  K.~Levin, V.~Lyzinski, and Y.~Qin, ``Statistical inference on random dot
  product graphs: a survey,'' \emph{The Journal of Machine Learning Research},
  vol.~18, no.~1, pp. 8393--8484, 2017.

\bibitem{chung2019graspy}
J.~Chung, B.~D. Pedigo, E.~W. Bridgeford, B.~K. Varjavand, H.~S. Helm, and
  J.~T. Vogelstein, ``Graspy: Graph statistics in python.'' \emph{Journal of
  Machine Learning Research}, vol.~20, no. 158, pp. 1--7, 2019.

\bibitem{vogelstein2019connectal}
J.~T. Vogelstein, E.~W. Bridgeford, B.~D. Pedigo, J.~Chung, K.~Levin, B.~Mensh,
  and C.~E. Priebe, ``Connectal coding: discovering the structures linking
  cognitive phenotypes to individual histories,'' \emph{Current opinion in
  neurobiology}, vol.~55, pp. 199--212, 2019.

\bibitem{ohyama2015multilevel}
T.~Ohyama, C.~M. Schneider-Mizell, R.~D. Fetter, J.~V. Aleman, R.~Franconville,
  M.~Rivera-Alba, B.~D. Mensh, K.~M. Branson, J.~H. Simpson, J.~W. Truman
  \emph{et~al.}, ``A multilevel multimodal circuit enhances action selection in
  drosophila,'' \emph{Nature}, vol. 520, no. 7549, pp. 633--639, 2015.

\bibitem{schneider2016quantitative}
C.~M. Schneider-Mizell, S.~Gerhard, M.~Longair, T.~Kazimiers, F.~Li, M.~F.
  Zwart, A.~Champion, F.~M. Midgley, R.~D. Fetter, S.~Saalfeld \emph{et~al.},
  ``Quantitative neuroanatomy for connectomics in drosophila,'' \emph{Elife},
  vol.~5, p. e12059, 2016.

\bibitem{mcinnes2018umap}
L.~McInnes, J.~Healy, and J.~Melville, ``Umap: Uniform manifold approximation
  and projection for dimension reduction,'' \emph{arXiv preprint
  arXiv:1802.03426}, 2018.

\bibitem{pedregosa2011scikit}
F.~Pedregosa, G.~Varoquaux, A.~Gramfort, V.~Michel, B.~Thirion, O.~Grisel,
  M.~Blondel, P.~Prettenhofer, R.~Weiss, V.~Dubourg \emph{et~al.},
  ``Scikit-learn: Machine learning in python,'' \emph{the Journal of machine
  Learning research}, vol.~12, pp. 2825--2830, 2011.

\end{thebibliography}
\end{document}